\pdfoutput=1

\documentclass[11pt, a4paper]{article}

\usepackage[acceptedWithA]{tacl2021v1} 
\usepackage{booktabs}

\usepackage{times}
\usepackage{latexsym}
\usepackage{amssymb}
\usepackage{graphicx}
\usepackage{microtype}
\usepackage{booktabs}
\usepackage{multirow}
\usepackage[utf8]{inputenc}
\usepackage[LFE,LAE,T1]{fontenc}
\usepackage[main=english,arabic]{babel}

\usepackage{hyperref}
\usepackage{makecell}
\usepackage{algorithm}
\usepackage{algpseudocode}
\usepackage{amsmath}

\usepackage{microtype}

\newcommand{\name}{MMC}

\title{Multilingual Coreference Resolution in Multiparty Dialogue}

\author{{Boyuan Zheng\textsuperscript{\rm 1} ~~~~~ Patrick Xia\textsuperscript{\rm 2}\thanks{~~Work done at JHU/HLTCOE.} ~~~~~ Mahsa Yarmohammadi\textsuperscript{\rm 1} ~~~~~ Benjamin Van Durme\textsuperscript{\rm 1}} \\
  \textsuperscript{1}~Johns Hopkins University \\
  \textsuperscript{2}~Microsoft Semantic Machines \\
  \texttt{\{bzheng12, mahsa, vandurme\}@jhu.edu}}

\begin{document}
\maketitle
\begin{abstract}
Existing multiparty dialogue datasets for entity coreference resolution are nascent, and many challenges are still unaddressed. We create a large-scale dataset, \textit{Multilingual Multiparty Coref} (\name), for this task based on TV transcripts.  Due to the availability of gold-quality  subtitles in multiple languages, we propose \textit{reusing} the annotations to create silver coreference resolution data in other languages (Chinese and Farsi) via annotation projection. %
On the gold (English) data, off-the-shelf models perform relatively poorly on \name, suggesting that \name~has broader coverage of multiparty coreference than prior datasets. On the silver data, we find success both using it for data augmentation and training from scratch, which effectively simulates the zero-shot cross-lingual setting. 
\end{abstract}

\section{Introduction}
\label{sec:intro}
\begin{figure}[t]
\includegraphics[width=1.0\linewidth]{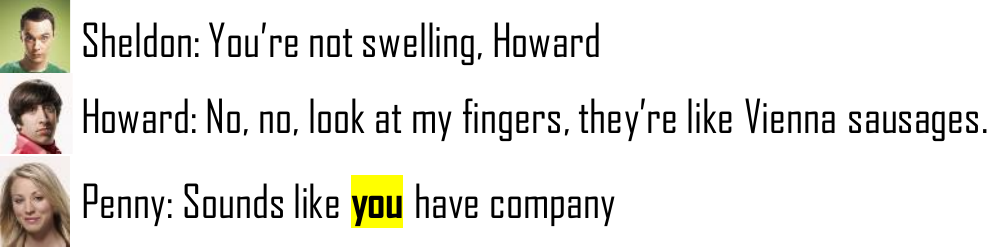}
\hrule
\includegraphics[width=1.0\linewidth]{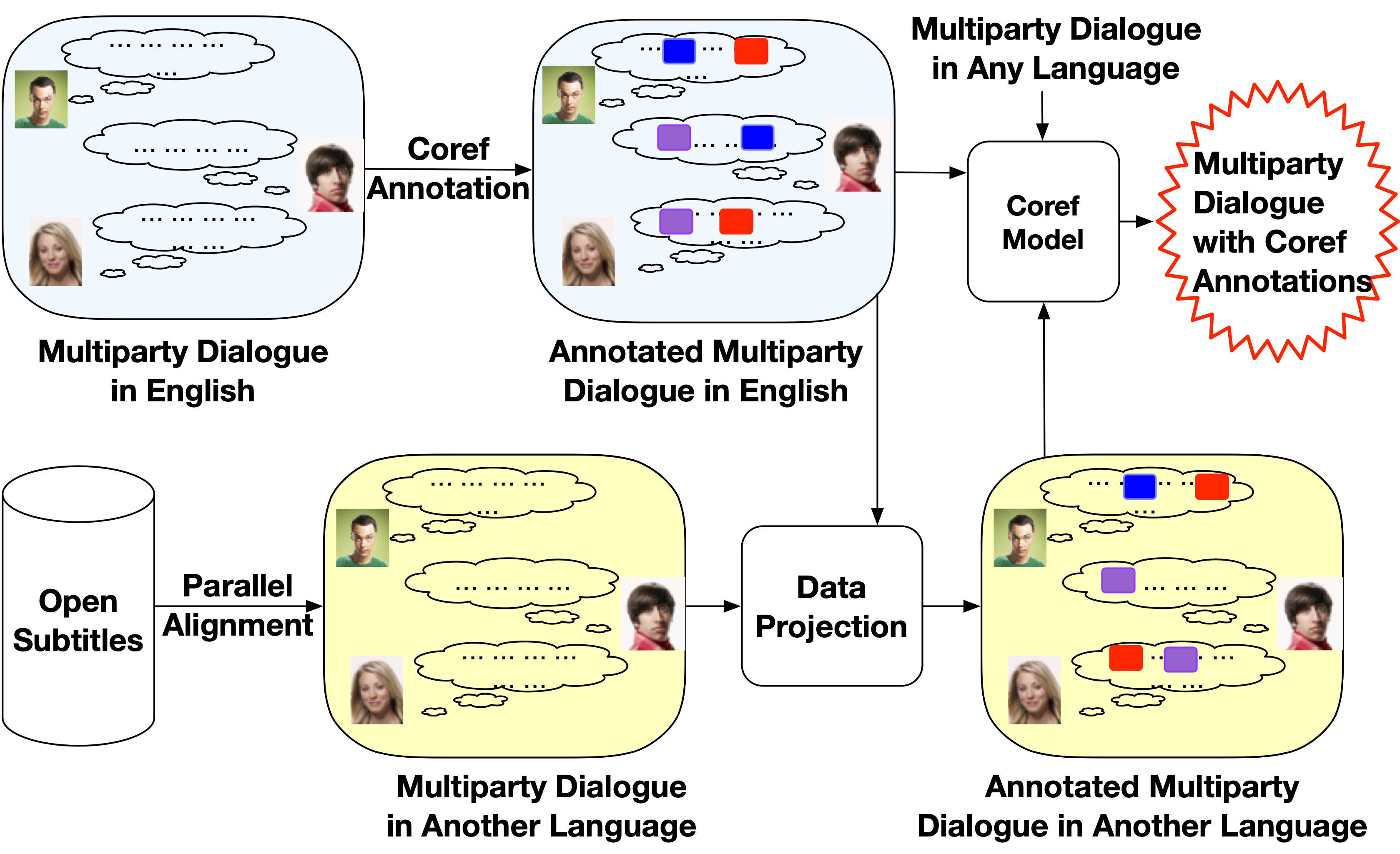}
\centering
\begin{small}
\caption{(\textit{Top}) An example of ambiguous coreference due to multi-person context: Penny's \emph{``you''} could refer to Sheldon, Howard, or both. %
(\textit{Bottom}) %
  Annotations can be projected to new languages, enabling model training beyond English.}
\label{fig-intro}
\end{small}
\end{figure}

Coreference resolution is a challenging aspect of understanding natural language dialogue  %
\cite{khosla-etal-2021-codi}. Many dialogue datasets are between two participants, even though there are distinct challenges that arise in the \textit{multiparty} setting with more than two speakers. Fig. \ref{fig-intro} shows how ``you'' could refer to any subset of the listeners of an utterance.
While there are some datasets on multiparty conversations from TV transcripts \cite{choi-chen-2018-semeval}, they only annotate \textit{people}, resulting in incomplete annotations across entity types. 
Moreover, these datasets are only limited to English, and works in dialogue coreference resolution in other languages are rare \cite{muzerelle-etal-2014-ancor}.

We introduce a new (entity) coreference resolution dataset focused on multiparty dialogue that supports experiments in multiple languages. We first annotate for coreference on the transcripts from two popular TV shows, in English. We then leverage existing gold subtitle translations~\cite{opensubtitle} in Chinese and Farsi to project our annotations, resulting in a multilingual corpus (Fig. \ref{fig-intro}).

Our experiments demonstrate that coreference resolution models trained on existing datasets are not robust to a shift to this domain. Further, we demonstrate that training on our projected annotations to non-English languages leads to improvements in non-English evaluation. Finally, we lay out an evaluation for zero-shot cross-lingual coreference resolution, requiring models to test on other languages with no in-language examples. We release over 1,200 scenes from TV shows with all annotations and related metadata in English, Chinese, and Farsi, which we call MMC: Multilingual Multiparty Coreference. %

\section{Motivation and Related Work}
\label{sec:background}
Many works on coreference resolution primarily study documents with a single author or speaker. OntoNotes \cite{weischedel2013ontonotes} is a widely used dataset that mostly consists of single-author documents, like newswire, while other datasets like PreCo \cite{chen-etal-2018-preco}, LitBank \cite{bamman-etal-2020-annotated}, WikiCoref \cite{ghaddar-langlais-2016-wikicoref} also consist of documents like books. Many recent modeling contributions also focus primarily on this setting and these datasets \cite{lee-etal-2017-end, lee-etal-2018-higher, xu-choi-2020-revealing, Bohnet2022Coreference} and some offload it to pretrained language models \cite{wu-etal-2020-corefqa, toshniwal-etal-2021-generalization} or ignore the speaker identity entirely \cite{xia-etal-2020-incremental} in an attempt to unify dialogue with non-dialogue domains. 

The dialogue domain is less studied because we lack a suitable dataset, even though these exist for other NLP tasks (Section \ref{related:multiparty-dialogue}). In addition to filling this gap, we also present a scalable solution for dataset creation in other languages, following related work in data projection methods (Section \ref{related:multilinguality}). The limitations of existing works motivate the creation of our dataset.

\subsection{Multiparty Conversations}
\label{related:multiparty-dialogue}
One of the focuses of this work is  \textit{multiparty} coreference resolution, which concerns coreference in conversational text with multiple participants. In particular, we are interested in conversations with more than two participants since this brings additional challenges not present in typical dialogue datasets. For example, in two-way conversations, ``you'' is typically deducible as the listener of an utterance. However, as shown in Fig. \ref{fig-intro}, ``you'' in multiparty conversations with more participants could refer to any of the participants present in the conversation. Additional challenges include using a third person pronoun to refer to one of the interlocutors and plural mentions (``we'', ``you all'') that refer to a subset of the participants in the conversation \cite{zhou-choi-2018-exist}.

Multiparty conversations are ubiquitous, especially in the form of spontaneous speech and dialogue. They have been used to study tasks like discourse parsing and  summarization \cite[i.a.]{afantenos-etal-2015-discourse, liu-chen-2021-improving, manuvinakurike-etal-2021-incremental}, and coreference resolution \cite{Walker1997StandardsFD, Jovanovic2005ACF, Frampton2009WhoI,choi-chen-2018-semeval}. %
Despite the breadth of domains and formality across all datasets, each multiparty dataset itself is narrowly focused, like meetings \cite{McCowan2005TheAM, hsueh-etal-2006-automatic}, board game play \cite{asher-etal-2016-discourse}, fantasy storytelling \cite{rameshkumar-bailey-2020-storytelling}, technical or persuasive online forums \cite{li-etal-2020-molweni, wang-etal-2019-persuasion} and sitcom transcripts \cite{choi-chen-2018-semeval, sang-etal-2022-tvshowguess}. %

\paragraph{Dialogue Coreference Resolution}

Coreference resolution in dialogue has recently reemerged as an area of research, with multiple datasets created and annotated for coreference resolution (\citet{li-etal-2016-large}, \citet{ khosla-etal-2021-codi}, more examples in \autoref{tab:datasets}) and the development of dialogue-specific models \cite{xu-choi-2021-adapted, kobayashi-etal-2021-neural, kim-etal-2021-pipeline}. The datasets can be broadly categorized into transcripts of spoken conversations (e.g. interviews), meeting notes, online discussions, and one-on-one goal-driven genres. \autoref{tab:datasets} shows that none of the datasets sufficiently covers spontaneous multiparty conversations. For the datasets that are multiparty, they are either incompletely annotated (Friends\textsubscript{CI} only annotates mentions referring to people), task-oriented (AMI), or discussion forums (ON-web, BOLT-DF). 
As a result, there are drawbacks to each of these datasets, like 
an expectation of formality (without the types of language found in spontaneous dialogue) or missing clarity on the listener or reader identities (e.g. missing usage of second person pronouns).  None of these datasets aim for exhaustive annotation on multiparty dialogue in spontaneous social interactions.

Friends\textsubscript{CI} \cite{choi-chen-2018-semeval} is the closest dataset to the goals of this work.\footnote{OntoNotes is also close, but each conversational genre has its own drawbacks.} Different from our goals, Friends\textsubscript{CI} is focused on \textit{character} linking instead of general entity coreference. While pronouns like ``you'' are annotated, other entities, like objects or locations, are not.
However, if we want to use coreference resolution models in  downstream systems for information extraction \cite{li-etal-2020-gaia} or dialogue understanding \cite{Rolih2018ApplyingCR, liu-etal-2021-coreference}, we need a dataset that aligns closer with multiparty spontaneous conversations. We contribute a large-scale and more exhaustively annotated dataset for multiparty coreference resolution.

\begin{table*}[ht]
    \centering
    \begin{small}
    \begin{tabular}{lccccc}
    \toprule
         Dataset & \makecell{Multiparty\\($>2$)} & \makecell{Exhaustive \\ Entities} & \makecell{Spontaneous} & \makecell{Clear \\Interlocutors} & \makecell{Multi-\\lingual} \\
         \midrule  
         TRAINS93 \cite{Byron1998ResolvingDA} & & \checkmark & & \checkmark \\
         Friends\textsubscript{CI} \cite{chen-choi-2016-character} & \checkmark &  & \checkmark & \checkmark \\
         ON (tc) \cite{weischedel2013ontonotes} & & & \checkmark & \checkmark& \checkmark \\
         ON (bc) \cite{weischedel2013ontonotes} & \checkmark &  &   &  \checkmark& \checkmark \\
         ON (wb) \cite{weischedel2013ontonotes} & \checkmark &  & \checkmark && \checkmark\\
         BOLT (DF) \cite{li-etal-2016-large} & \checkmark & & \checkmark & \\
         BOLT (SMS, CTS) \cite{li-etal-2016-large} & & & \checkmark & \checkmark \\ 
         AMI\textsuperscript{C} \cite{McCowan2005TheAM} & \checkmark & \checkmark & & \checkmark \\ 
         Persuasion\textsuperscript{C} \cite{wang-etal-2019-persuasion} & &  \checkmark & & \checkmark \\
         Switchboard\textsuperscript{C} \cite{stolcke-etal-2000-dialogue} & & \checkmark & \checkmark & \checkmark\\ 
         LIGHT\textsuperscript{C} \cite{urbanek-etal-2019-learning} & & \checkmark & & \checkmark \\
         \midrule
         \name~(Our work)& \checkmark & \checkmark & \checkmark & \checkmark & \checkmark\\
    \bottomrule
    \end{tabular}
    \caption{Examples of dialogue coreference datasets. Nothing to our knowledge satisfies our desire for modeling spontaneous multiparty conversations. Additionally, parallel data is available for \name, which enables exploration in non-English languages. Superscript \textsuperscript{C} indicates that they were additionally annotated by \citet{khosla-etal-2021-codi}. OntoNotes (ON) is divided by genre.}
    \label{tab:datasets}
    \end{small}
\end{table*}

\subsection{Multilinguality}
\label{related:multilinguality}
\paragraph{Coreference Resolution}
Coreference resolution models are typically developed for a single language, and while there is some prior work on cross-lingual and multilingual models \cite{xia-van-durme-2021-moving}, these methods still require some data in the desired language for best performance. While there are coreference resolution datasets in many languages \cite{weischedel2013ontonotes, recasens-etal-2010-semeval}, they are often limited and expensive to annotate from scratch for each new language. We take a step towards a more general solution for building coreference resolution models from scratch in (almost) any language. By collecting and annotating data that already exists in a highly parallel corpus, we suggest a different approach to expensive in-language annotation: data projection.

\paragraph{Data Projection}

Using annotations in English to create data in a target language has been useful for tasks such as semantic role labeling \cite{akbik-etal-2015-generating,aminian-etal-2019-cross}, information extraction \cite{riloff-etal-2002-inducing}, POS tagging \cite{yarowsky-ngai-2001-inducing}, and dependency parsing \cite{ozaki-etal-2021-project}. Previous works find improvements when training on a mixture of gold source language data and projected silver target language data in cross-lingual tasks such as semantic role labeling \cite{fei-etal-2020-cross,daza-frank-2020-x} and information extraction \cite{yarmohammadi-etal-2021-everything}.
The intuition of using both gold and projected silver data is to allow the model to see high-quality gold data as well as data with target language statistics. In this work, we extend projection to coreference resolution both for creating a model without in-language data and for augmenting existing annotations.

\section{Multilingual Multiparty Dialogue Coreference Dataset}
\label{sec:dataset}
In this section, we present our multilingual multiparty coreference (\name) dataset\footnote{This dataset is released under the Apache License.}, including the construction process of data alignment and filtering, annotation, and projection.\footnote{Dataset, software, and annotation infrastructure used in this work are available at: \url{https://github.com/boyuanzheng010/mmc}. } Core to our contribution is the choice of a multiparty dataset that \textit{already has gold translations} and prioritizing multilinguality throughout the data collection process. %

\begin{figure*}[t]
\includegraphics[width=\textwidth]{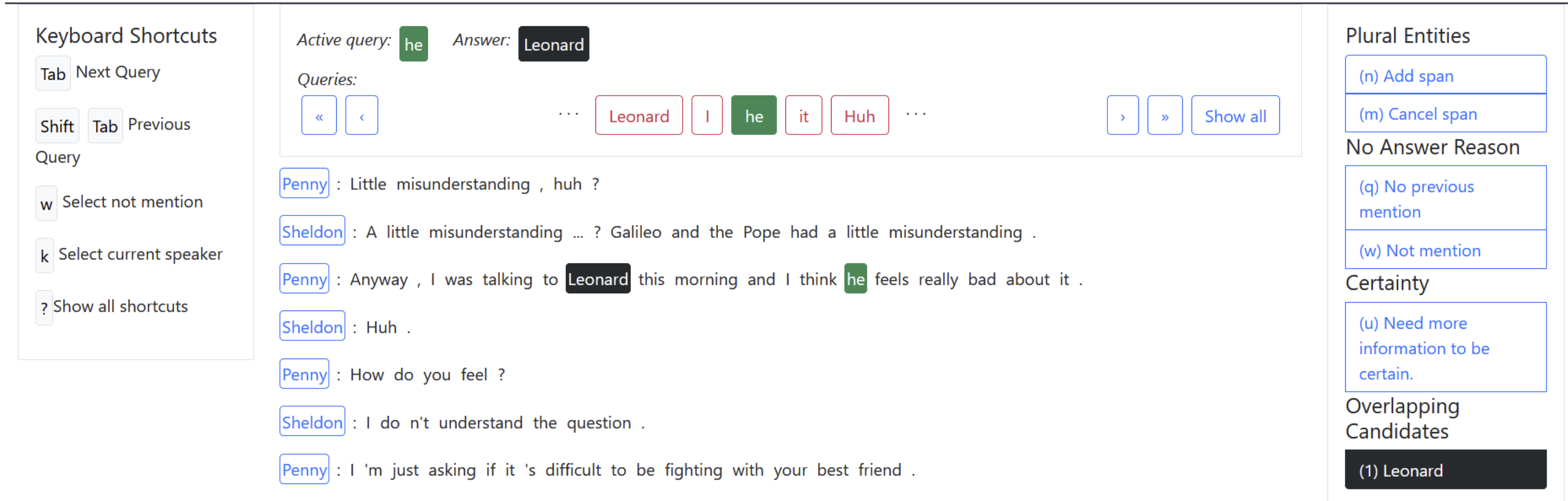}
\centering
\caption{This figure illustrates the annotation interface. Given a set of proposed markables (``queries''), users highlight the best antecedent \textit{or speaker} that the markable refers to or select ``no previous mention'' or ``not a mention.'' Plural entities and uncertainty due to missing context can also be annotated.}
\label{fig:annotation-ui}
\end{figure*}

\subsection{Parallel Dialogue Corpus}
We construct a parallel corpus of multiparty dialogue by aligning the English transcripts from TV shows and parallel subtitles from the OpenSubtitles corpus \cite{tiedemann-2012-parallel,  lison-tiedemann-2016-opensubtitles2016}, a sentence-aligned parallel corpus widely used in machine translation.\footnote{\url{https://www.opensubtitles.org/}} 

TV sitcoms are an ideal target for meeting our criteria for a spontaneous multiparty genre, as they contain rich multiparty dialogues, multiple references to interlocutors, and spontaneous utterances.\footnote{While transcripts are pre-written, they are written to mimic spontaneous speech.} We select \textit{Friends} and \textit{The Big Bang Theory} (TBBT) because there is prior work in preprocessing and speaker identification for the transcripts of these shows  \cite{tvd, choi-chen-2018-semeval,  sang-etal-2022-tvshowguess}. 

We align the available data with that from two languages distant from English: Chinese and Farsi (Section \ref{sec:alignment}). Due to missing episodes and alignments for some languages, the final three-way aligned corpus is an intersection of what is available in all three languages, and empty or clearly misaligned scenes are removed (Table \ref{tab:filter}). %

\addtolength{\tabcolsep}{-1.5pt}  
\begin{table}[]
\begin{center}
\begin{small}
\begin{tabular}{ccccccc}
\toprule
\multicolumn{2}{l}{\textbf{TV Show}} & \textbf{Ep.} & \textbf{Scenes} & \textbf{Utter.} & \textbf{Speakers}  \\

\midrule
\multirow{3}{*}{TBBT} & 2-way &184 &2,212 &40,883 & 432 \\
& 3-way & 88 & 1,086 & 19,773 & 249   \\
& final & 88 & 979 & 18,350& 191\\
\midrule
\multirow{3}{*}{Friends} & 2-way & 28 & 368 & 7,113 & 146 \\ 
& 3-way & 21 & 270 & 5,226 & 114 \\
& final & 21 & 243& 4,614& 104\\
\bottomrule
\end{tabular}
\caption{Source data statistics (episodes, scenes, utterances, unique speakers) before and after filtering for three-way alignable episodes. \textit{2-way} contains the union of two-way alignable episodes, while \textit{3-way} contains the intersection, i.e. three-way alignable episodes. After a \textit{final} filtering step, there are 1,222 scenes in total.}
\label{tab:filter}
\end{small}
\end{center}
\end{table}
\addtolength{\tabcolsep}{1.5pt}

\subsection{English Coreference Annotation}
\label{sec:english}

We automatically create an initial set of proposed markable mentions, aiming for high recall. Like prior work \cite{pradhan-etal-2012-conll, poesio-etal-2018-anaphora, bamman-etal-2020-annotated}, for consistent annotation, these markables are then considered for coreference linking. 
We mainly follow the annotation process of OntoNotes 5.0~\cite{weischedel2013ontonotes}.\footnote{Details are in the annotation interface instructions.}
However, we make some simplifications that are easier to understand for crowdworkers, roughly following those made by \citet{chen-etal-2018-preco}. Unlike OntoNotes, we do not consider verbs and verb phrases as markable. Entities mentioned once (\textit{singletons}) are annotated. Also, non-proper modifiers can be coreferred with generic mentions, and subspans can be coreferred with the whole span.

\paragraph{Markable Mention Proposal}

We ensemble predictions from the Berkeley parser with T5-Large \cite{berkeley-parser, 2020t5} and RoBERTa-based \cite{roberta} spaCy\footnote{We use the en\_core\_web\_md-3.2.0 model.} to detect nouns, noun phrases, and pronouns. These constitute our proposed markable mention spans.

\paragraph{Interface}

Our annotation interface (Fig. \ref{fig:annotation-ui}) is derived from that of \citet{yuan-etal-2022-adapting}. The interface simplifies coreference annotation to selecting \textit{any} antecedent for each query span (proposed markable) found by the parser. For consistency, the interface encourages users to select proposed markables, although they can also add a new antecedent mention if it is not among those proposed by the parser. %
They can also label a markable span as not a mention. Coreference clusters are formed by taking the transitive closure after annotation. 

We make several modifications to the interface to annotate coreference more completely and in the dialogue setting. These include permitting the selection of speakers, mentions of arbitrary size for plural mentions, and an indication of uncertainty (e.g. without further context, the example in Fig. \ref{fig-intro} requires audiovisual cues). While the annotation of plural mentions and uncertainty labels are not used in this work, we hope they enable future studies.

\paragraph{Pilot Annotation}

We sampled three scenes of differing lengths from the training set for a qualification study. For these scenes, we adjudicated annotations from four experts as the gold labels. Then, we invited annotators from Amazon Mechanical Turk to participate and receive feedback on their errors. Nine high-scoring annotators on the pilot\footnote{The lowest Cohen's Kappa score is 0.6549} (>80 MUC score) were selected for annotating the full training set. We paid USD \$7 for each pilot study, which could be completed in 25-35 minutes, including reading the instructions.

\paragraph{Full Annotation}

For the training set, the scenes were batched into roughly 250 proposed markables each. We paid \$4 per batch (expected \$12/hour) for each of the nine high-scoring annotators. Each of the scenes was annotated once, although we inspected these annotations to ensure they were nontrivial (i.e. not all-blank or all-singletons).

For the dev and test splits, three professional data specialists,\footnote{These are in-house data annotators with linguistics background who are trained by and correspond with the authors during the annotation process.} in consultation with the authors, annotated the documents with two-way redundancy.
After reviewing common error and disagreement types with the authors, one of the specialists performed adjudication of the disagreements (described in \autoref{sec:merge-two-way}).
Following several prior works \cite{weischedel2013ontonotes, chen-etal-2018-preco, Toshniwal2020LearningTI}, we adopt MUC score as an approximate measure of agreement between annotators. The average MUC score of each annotator to the adjudicated clusters is 86.1. This agreement score is comparable to reported scores in widely used datasets: OntoNotes (89.60), PreCo (85.30). The inter-annotator MUC agreement score on this combined split is 80.3 and the inter-annotator CoNLL-F1 score is 81.55. The Cohen's Kappa score is 0.7911, which is interpreted as "substantial agreement." Note that the high agreement can be partially attributed to the agreement over non-mentions and starts of coreference chains.

7.2\%, 8.8\%, and 10.0\% of the clusters in training, dev, and test splits contain plural mentions. Meanwhile, 0.4\%, 1.4\%, and 1.6\% of the mentions are marked as ``uncertain.'' The specialists working on the dev and test sets were more likely to mark an annotation as uncertain than crowdworkers.

\begin{figure*}[t]
\includegraphics[width=\linewidth]{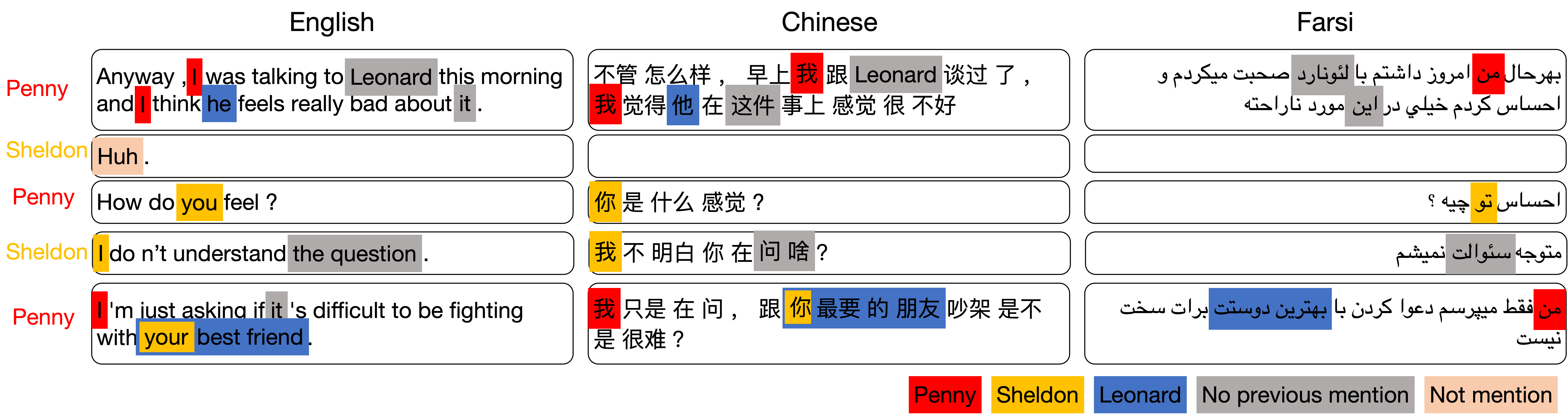}
\centering
\caption{Example utterances with gold English and projected Chinese and Farsi coreference annotations.}
\label{fig:projection-example}
\end{figure*}

\subsection{Silver Data via Annotation Projection}
\label{sec:alignment}
Data projection transfers span-level annotations in a source language to a target language via word-to-word alignments in a fast, fully-automatic way. The projected (\emph{silver}) target data can be used directly or combined with gold data as data augmentation. %

\paragraph{Alignment} We need to align English (source side) mention \textit{spans} to Chinese or Farsi (target side) text spans. Our cleaned dataset contains utterance-level aligned English to Chinese and Farsi text. Using automatic tools, we obtain more fine-grained (word-level) alignments, and project source spans to target spans according to these alignments. %
For multi-token spans, the target is a contiguous span containing all aligned tokens from the source span.  

We use Awesome-align~\citep{dou-neubig-2021-word}, a contextualized embedding-based word aligner\footnote{\url{https://github.com/neulab/awesome-align}} that extracts word alignments based on token embedding similarities. 
We fine-tune the underlying XLM-R encoder on around two million parallel sentences from the OSCAR corpus \cite{2022arXiv220106642A}. We further fine-tune on Farsi-English gold alignments %
by \citet{Tavakoli-2014} and the GALE Chinese–English %
gold alignments  \cite{Li2015GALECP}. See \autoref{sec:appendix-awesome} for dataset statistics and fine-tuning hyperparameters. By projecting the annotated English mentions to the target side, the entity clusters associated with each mention are also implicitly projected.

Some coreference annotations are not transferred to the target language side either due to empty subtitles in our cleaned data
or erroneous automatic word alignment and projection of the source text span. We refer to such cases as \emph{null projections}.

Fig. \ref{fig:projection-example} shows parallel utterances with their gold English and projected Chinese and Farsi annotations. Some short English utterances do not have counterparts, such as the second utterance (``Huh"). Chinese and Farsi annotations are also a subset of English annotations due to null projections. For example, the English mention ``it" in the last utterance is missing in the target transcripts, so this span's annotation is missing in the projected data.

While there are the same number of episodes in the English and projected data, the number of scenes, mentions, and clusters in the projected data are smaller due to missing scenes or null projections. We see around 30\% (Zh) and 40\% (Fa) drop in aligned mentions (Table \ref{tab:silver-data}). %

\addtolength{\tabcolsep}{-1.5pt}  
\begin{table}[]
\small
\begin{center}
\begin{tabular}{llrrrr}
\toprule
\multicolumn{2}{l}{\textbf{Lang. (split)}} & \textbf{Scenes} & \textbf{Utter.} & \textbf{Ment.} & \textbf{Clusters}  \\
\midrule
\multirow{3}{*}{En} & train & 955 & 18,477 & 60,997 & 25,349\\
 & dev & 134 & 2,578 & 10,079 & 4,804 \\
 & test &  133 & 1,909 & 7,958 & 3,808  \\
 \midrule
\multirow{4}{*}{Zh} & train & 948 & 14,467 & 42,234 & 20,251   \\
 & dev &  134 & 2,146 & 7,922 & 4,193   \\
 & test\textsubscript{silver} & 133 & 1,611 & 5,977 & 3,106  \\
 & test\textsubscript{correct} & 133 & 1,611 & 5,642 & 2,934  \\
 \midrule
 \multirow{3}{*}{Fa} & train & 948 &  14,357 &  36,415 &  20,063 \\
 & dev & 120 & 1,983 & 5,887 & 3,566  \\
 & test\textsubscript{silver} & 133 & 1,612 & 4,894 & 3,021   \\
 & test\textsubscript{correct} & 133 & 1,612 & 6,053 & 3,169  \\
\bottomrule
\end{tabular}
\caption{\name~statistics. English (En) is manually annotated while Chinese (Zh) and Farsi (Fa) are projected. Since all episodes are three-way parallel, the splits for each language contain the same scenes (some empty scenes are omitted).} 
\label{tab:silver-data}
\end{center}
\end{table}
\addtolength{\tabcolsep}{1.5pt}

\paragraph{Alignment Correction} We conducted alignment annotation for both English-Chinese and English-Farsi utterances to collect alignment corrections for the Chinese and Farsi test set with four Chinese speakers and three Farsi speakers. For each language pair, we presented the user with the utterance in each language and one of the English spans highlighted. On the target language side, the prediction by the projection model is displayed. The user makes corrections to the automatic alignments if necessary. This is conducted via the TASA interface\footnote{\url{https://github.com/hltcoe/tasa}} \cite{stengel-eskin-etal-2019-discriminative}. 
These corrected annotations serve as the test set for both Chinese and Farsi. 1,904 (24.81\%) projections are corrected in Chinese and 2,485 (32.26\%) projections in Farsi. There are three types of corrections: addition, deletion, and modification, shown in Table \ref{tab:correction_stats}. For \textit{addition}, a mention boundary is added for a null projection. For \textit{deletion}, the predicted projection is discarded. \textit{Modification} is where the predicted mention boundaries are modified. 

Chinese and Farsi are \textit{pro-drop} languages. Most of the \textit{addition} operations are related to pronouns, where the target is corrected from an empty string to the location of the trace of the pronoun (in Chinese) or the implied pronoun affix (in Farsi). In the \textit{modification} operation, a small amount of target mentions are also corrected to an empty string. This resulted in 401 and 823 additional pronoun mentions in Chinese and Farsi respectively.

\begin{table}[t]
\begin{center}
\begin{small}
\begin{tabular}{l|ccc}
\toprule
 Language  & Addition & Deletion & Modification \\
\midrule
 Chinese &609 (340)&	441 &916 (61)\\
 Farsi&	1,504 (739)&	187& 794 (84)\\
\bottomrule
\end{tabular}
\caption{Corrections in Chinese and Farsi test sets. The number in brackets is the number of dropped pronouns that are recovered.}
\label{tab:correction_stats}
\end{small}
\end{center}
\end{table}

\paragraph{Dataset Statistics} \name~ contains about 101 hours of episodes, resulting in 323,627 English words, 226,045 Chinese words, and 258,244 Farsi words. Table \ref{tab:silver-data} shows the final statistics of our three-way aligned, multiparty coreference corpus. This dataset is used for the remainder of the paper. To summarize, English dev and test data are two-way annotated followed by adjudication; English train is one-way annotated; and Chinese and Farsi are automatically derived via projection, but both Chinese and Farsi test alignments are corrected. %

\addtolength{\tabcolsep}{-1.5pt} 
\begin{table}[h]
\small
\begin{center}
\begin{tabular}{lllrrrr}
\toprule
Dataset & \# Speakers & Lang. & Train & Dev. & Test\\
\midrule
\multirow[t]{4}{*}{ON\textsuperscript{En}} &  all & En & 2,802 & 343 & 348 \\
& $\leq 1$ & En &2,321 &260 &263 \\
& 2 & En & 255&43 &47 \\
& $>2$ & En &226 &40 &38 \\
Friends\textsubscript{CI} & all & En &1,041 &130 &130 \\
\midrule 
\multirow[t]{4}{*}{ON\textsuperscript{Zh}} & all & Zh & 1,810 & 252 & 218 \\
& $\leq 1$ & Zh & 1,343& 165& 159\\
& 2 & Zh & 173&22 &18 \\
& $>2$ & Zh & 294&65 &41 \\
\bottomrule
\end{tabular}
\caption{Statistics for additional datasets used in this work. OntoNotes has a Chinese split; we are not aware of other Farsi coreference datasets.} 
\label{tab:training-stats}
\end{center}
\end{table}
\addtolength{\tabcolsep}{1.5pt}

\label{sec:monolingual}
\addtolength{\tabcolsep}{-1.25pt} 
\begin{table*}[t]
\begin{center}
\begin{small}
\begin{tabular}{ll|ccccccccc}
\toprule
 \textbf{Train} & &&&&&\textbf{Test}  \\
  &&ON &Friends\textsubscript{CI}& \name&ON\textsubscript{$\leq1$} &ON\textsubscript{2} &ON\textsubscript{$>2$}&\name\textsubscript{Friends}& \name\textsubscript{TBBT}  \\
\midrule
  \multirow{6}{*}{En}&\citet{xu-choi-2020-revealing} & \textbf{79.96} &54.41& 50.97& \textbf{81.56}& \textbf{78.16}&\textbf{75.67}&48.49&51.60\\
  &ON\textsuperscript{En} &78.80 &54.60 &53.41 &80.41 &76.90 &74.59 & 49.30& 50.14& \\
  &Friends\textsubscript{CI} & 30.91 &\textbf{71.19}& 46.64&24.52&43.32&36.67&49.99&45.76 \\
  &\name &46.53&48.10 & 70.71&42.97&54.45&48.62&67.02&71.61\\
  &\name~(XLM-R)&35.08&44.52&69.63&32.76&41.17&35.45&67.08&70.25\\
  &ON\textsuperscript{En}+\name &73.62 &48.42 &72.01&74.94&72.58&69.66&69.26&72.67\\
  &ON\textsuperscript{En}$  \rightarrow$\name &65.72 & 47.40 &\textbf{75.87}&67.00&65.06&61.29&\textbf{73.01}&\textbf{76.58}\\
 \midrule
\multirow{4}{*}{Zh}\ & ON\textsuperscript{Zh} & \textbf{65.52}&--&36.56&\textbf{66.59}&65.62&60.89&36.58&36.56\\
 & \name & 23.71&--&47.89&14.72&47.12&38.92&\textbf{43.33}&49.01\\
 & ON\textsuperscript{Zh}+\name & 64.19&--& 47.37&63.79&\textbf{68.87}&\textbf{63.24}&41.79&48.70\\
 & ON\textsuperscript{Zh}$\rightarrow$\name&37.52&-- & \textbf{48.65}&33.32 &51.51& 44.74&42.56&\textbf{50.01}\\
\bottomrule
\end{tabular}
\caption{F1(\%) scores of models trained on a combination of different datasets for English and Chinese. All English models except MMC (XLM-R) use SpanBERT-Large as the encoder, while MMC (XLM-R) and Chinese models use XLM-R-base as the encoder.}%
\label{tab:monolingual}
\end{small}
\end{center}
\end{table*}
\addtolength{\tabcolsep}{1.25pt} 

\section{Methods}
\subsection{Model}
\label{sec:model}
For all experiments, we use the higher-order inference (HOI) coreference resolution model \cite{xu-choi-2020-revealing}, modified slightly to predict singleton clusters \cite{xu-choi-2021-adapted}. Given a document, HOI encodes texts with an encoder and enumerates all possible spans to detect mentions. These spans are scored by a mention detector, which prunes the spans to a small list of candidate mentions. The candidate mentions are scored pairwise, corresponding to the likelihood of being coreferring, and the resulting scores are used in clustering. While mentions can be linked to their top-scoring antecedent, higher-order inference goes further and ensures high agreement between all mentions in a cluster by making additional passes.
Singletons can be predicted when a high-scoring (via the mention detector) mention only has low-scoring (via the pairwise scorer) candidate antecedents. For English-only experiments, SpanBERT-large~\cite{joshi-etal-2020-spanbert} is used as the encoder while for the other experiments XLM-R-base~\cite{conneau-etal-2020-unsupervised} is used. More hyperparameter details are in Appendix \ref{sec:appendix-hyperparameters}.\footnote{Code to run experiments: \url{https://github.com/boyuanzheng010/mmc}}

\subsection{Noise-tolerant Mention Loss}
\label{sec:noise_loss}
The loss function used by \citet{xu-choi-2021-adapted} consists of a cluster loss, $L_c$,\footnote{$L_c$ is based on the pairwise scorer. It is the (marginal) log-likelihood of all correct antecedent mentions for a single mention, which is provided by the gold clusters. We do not modify it in this work.} typically used for coreference resolution \cite{lee-etal-2017-end, joshi-etal-2020-spanbert, xu-choi-2020-revealing} and a binary cross-entropy mention detection loss, $L_m$, used to better predict singleton losses. %

Compared to the two-way annotated and merged dev/test set, the one-way annotated train set is more likely to be subject to annotator biases, leading to noise in the train set. These inconsistencies are further exacerbated when projected to silver data, leading to a low recall of mentions in training, as evidenced by the number of ``additions'' in \autoref{tab:correction_stats}.

To address this noise, we propose a modification of $L_m$ to downweight negative labels. Following the notation from \citet{xu-choi-2021-adapted}, let $\Psi_i^+$ be the set of gold candidate mentions and $\Psi_i^-$ be the remainder of the candidate spans. Applying a hyperparameter $\tau \in [0, 1]$, we can rewrite binary cross-entropy loss, $L_m^\tau$, as
$$\sum_{x_i\in \Psi^+}{\log(P(x_i))} + \tau \sum_{x_i \in \Psi^-}{\log{(1-P(x_i)}})$$ where $x_i$ is a candidate span and $P(x_i)$ is the output of the mention scorer. %
Following \citet{xu-choi-2021-adapted}, the mention loss is also weighted in the final loss, $L= L_c + \alpha_m L_m^{\tau}.$

\subsection{Data}
\label{sec:data}
We evaluate the performance of models across three datasets: \name, OntoNotes \cite{pradhan-etal-2012-conll}, and Friends\textsubscript{CI}~\cite{choi-chen-2018-semeval}. OntoNotes is a collection of documents spanning multiple domains, some of which include multiparty conversations or dialogues, like weblogs, telephone conversations, and broadcast news (interviews). Furthermore, OntoNotes is available in English and Chinese. Friends\textsubscript{CI} is a collection of annotations on TV transcripts from \textit{Friends}, including entity linking where character entities are linked. 
As the focus of this work is on multiparty conversations, we further separate OntoNotes into documents with 0 or 1 (ON\textsubscript{$\leq1$}), 2 (ON\textsubscript{2}), or more than two (ON\textsubscript{$>2$}) speakers/authors for evaluation. We didn't include split antecedents and drop-pronouns in the experiment, since the baseline model doesn't support predicting them. The statistics of datasets used in our experiments are in Table~\ref{tab:training-stats}.

\subsection{Evaluation} 
\label{sec:evaluation}
We use the average of MUC \cite{vilain-etal-1995-model}, B\textsuperscript{3} \cite{Bagga98algorithmsfor}, and CEAF$_{\phi_4}$ \cite{luo-2005-coreference}), which is also used for OntoNotes. Furthermore, each model is trained three times and the average test score (CoNLL\textsubscript{F1}) is reported.

\section{Experiments and Results}

First, we highlight the differences of \name~in contrast to Friends\textsubscript{CI} and OntoNotes. To do so, we train an off-the-shelf model on the three datasets. Additionally, we establish monolingual baselines for all three languages. Finally, we explore the cross- and multi-lingual settings to validate the recipe of using data projection for coreference resolution. %

\subsection{Monolingual Models}
Table \ref{tab:monolingual} shows the performance of several monolingual models. They highlight that models trained on other datasets (ON, Friends\textsubscript{CI}) perform substantially worse than models trained in-domain (on \name). Additionally, we find that both combining datasets and using continued training from OntoNotes \cite{xia-van-durme-2021-moving} can be effective for further improving model performance: for English, this leads to gains of 2.7 F1 points (combining) and 5.2 F1 points (continued training), and continued training is also effective in Chinese. 

Notably, combining the Chinese datasets yields the best scores on dialogues (ON\textsubscript{2}, ON\textsubscript{$>2$}) in OntoNotes. This highlights the utility of the \textit{silver} MMC data as a resource for augmenting preexisting in-domain data. Combining data is less helpful for English than Chinese possibly because there is more training data in ON\textsuperscript{En} than ON\textsuperscript{Zh}, making the Chinese data augmentation more useful. The baselines for ON\textsuperscript{Zh} may also be less optimized by prior work than models for ON\textsuperscript{En}.

\subsection{Cross-lingual and Multilingual Models}
\label{sec:projected}

\addtolength{\tabcolsep}{-1.5pt} 
\begin{table}[t]
\begin{center}
\begin{small}
\begin{tabular}{l|ccc}
\toprule
 \textbf{Train} & \multicolumn{3}{c}{\textbf{Test}}  \\
  &\name-En &\name-Fa& \name-Zh \\
\midrule
  Head Lemma &17.30&-&12.68\\
  \name-En & \textbf{69.63} & 22.91& 37.18 \\
  \name-Fa &55.99 &\textbf{35.00}& 36.85 \\
  \name-Zh & 45.91& 17.09&\textbf{47.89} \\
  \name-En-Zh-Fa &69.57 &33.49 &45.54  \\
\bottomrule
\end{tabular}
 \caption{Performance of models trained on datasets of different languages (English, Farsi, and Chinese) and the combination of all three of them. All four models use XLM-R-base as the encoder. 
 }
\label{tab:cross-lingual}
\end{small}
\end{center}
\end{table}
\addtolength{\tabcolsep}{1.5pt} 

Next, we demonstrate the ability of the silver data in Chinese and Farsi to contribute towards creating a model with no in-language coreference resolution data. While Chinese and Farsi are the two languages we choose to study in this work, parallel subtitles for the TV Shows in \name~are available in at least 60 other languages and can be used similarly, given a projection model.\footnote{We assume that a language contains parallel subtitles if we find alignable episodes within the bitext between English and that language. This count is approximate and is not exhaustive.}

\paragraph{Simple Baseline}
We adopt a simple head lemma match baseline to determine a lower bound for each language if we did not have any training data. %
We first find the NP constituencies as candidate mentions derived from off-the-shelf constituency parsers. We adopt the Berkeley parser with T5-Large \cite{berkeley-parser, 2020t5} for English and multilingual self-attention parser~\cite{Kitaev2019MultilingualCP} with Chinese ELECTRA-180G-large~\cite{Cui2020RevisitingPM} for Chinese.
For Farsi, we adopted the constituency parser in DadmaTools~\cite{Etezadi2022DadmaToolsNL}.\footnote{We use spaCy for English (en\_core\_web\_sm-3.4.0) and Chinese (zh\_core\_web\_sm-3.4.0), and Hazm toolkit~(\url{https://www.roshan-ai.ir/hazm/}) for Farsi.} However, we were not confident in the Farsi parser quality (under 5 CoNLL\textsubscript{F1} when evaluated on Farsi MMC), and could not find another widely used constituency parser for Farsi, so we omit Farsi in our results. To predict the clusters, we extract and lemmatize the head word for each mention. We link any two mentions that have the same head word lemma.

\paragraph{Cross-lingual Transfer}
We evaluate the monolingual XLM-R models for English, Chinese, and Farsi on each of the languages, i.e. ``test'' for English, ``test\textsubscript{correct}'' for Chinese, and ``test\textsubscript{silver}'' for Farsi. This effectively evaluates the zero-shot ability for the other two languages. 

Table \ref{tab:cross-lingual} shows that models trained on English data or silver projected data in Farsi and Chinese can achieve reasonable performance on the test set of its own language. Models trained on projected silver data in Farsi and Chinese achieve the best performance among their own test set compared with zero-shot performance of models trained in another language. Consequently, this implies that a recipe of projecting a coreference resolution dataset to another language and using that data to train from scratch outperforms naive zero-shot transfer via multilingual encoders.

\paragraph{Multilingual Models}

We combine the training data of three languages and train multilingual models. %
Table \ref{tab:cross-lingual} shows that these multilingual models achieve slightly to moderately worse performance on each test set compared to their monolingual counterparts. This contrasts with prior work \cite{fei-etal-2020-cross,daza-frank-2020-x,yarmohammadi-etal-2021-everything} that finds benefits to using silver data. The poorer performance of the multilingual model could be due to using the same set of hyperparameters for all three languages. While it does not surpass the monolingual models, it enjoys the benefits of being more parameter efficient.%

\subsection{Noise-tolerant Loss Results}

\begin{table}[t]
\begin{center}
\begin{small}
\begin{tabular}{l|ccc}
\toprule
 $L_m$ & \multicolumn{3}{c}{Language}  \\
  &\name-En &\name-Fa& \name-Zh \\
\midrule
  $\tau = 1$ & 69.63 & 35.00& 47.76 \\
  $\tau = 0$ &59.49 &31.32& 37.30 \\
  weighted &\textbf{72.58} & \textbf{37.05}&\textbf{49.56} \\
\bottomrule
\end{tabular}
\caption{Test set performance of models trained with different $\tau$. $\tau=1$ is the regular binary cross-entropy mention loss reported earlier in the paper. $\tau$ is chosen according to a grid search (Sec. \ref{sec:noise_loss}).}
\label{tab:noise-tolerant-loss}
\end{small}
\end{center}
\end{table}

Table~\ref{tab:noise-tolerant-loss} shows model performance using our modified loss. We find some benefits to downweighting negatively labeled spans, obtaining 1-3 points improvement compared to the original loss across all three languages.\footnote{With the weighting, XLM-R-base can outperform SpanBERT large in English} Thus, \name~could also enable exploration into additional \textit{modeling} questions around the use of projected and noisy annotations.

\section{Analysis}

We analyze our modeling results in relation to our original motivation. %
First, we explore differences between datasets (Section~\ref{section:analysis-dataset}), the number of speakers (Section~\ref{sec:analysis-speakers}), and overfitting (Section~\ref{sec:speaker}). For the data construction, we analyze the alignment corrections process (Section~\ref{sec:alignment-correction}) and compare recipes for annotation projection (Section~\ref{sec:annotation-from-scratch}).

\subsection{Comparison of Datasets}
\label{section:analysis-dataset}
Since Friends\textsubscript{CI} is also based on TV Shows (\textit{Friends}) and its dataset overlaps with \name, we would expect a model trained on Friends\textsubscript{CI} to perform well on \name. Instead, we find that its performance is over 23 F1 points worse. The main difference between Friends\textsubscript{CI} and \name~is that Friends\textsubscript{CI} only annotates characters instead of all possible mentions, and therefore there are fewer mentions per document in Friends\textsubscript{CI} than in \name. A closer inspection of the precision and recall appears to validate this hypothesis, as the macro precision (across the three metrics) is 65.8\% compared to a recall of 37.5\%. This is also evident in the mention span precision and recall, where a model trained on Friends\textsubscript{CI} scores 91.5\% precision but only 50.3\% on recall. We see the same trend for OntoNotes: high precision and low recall both on the coreference metrics and on mention boundaries. %

\subsection{Number of Speakers}
\label{sec:analysis-speakers}

Table \ref{tab:monolingual} also shows that in OntoNotes, models perform poorer on documents with more speakers. However, this is not the case with both Friends\textsubscript{CI} and \name, which perform best on two-person dialogues.\footnote{Many documents in ON\textsubscript{$\leq1$} are written documents, not dialogues, and therefore out-of-domain for Friends\textsubscript{CI} and \name.} Nonetheless, the drop in performance from ON\textsubscript{2} to ON\textsubscript{$>2$} highlights the additional difficulty of multiparty dialogue documents (in OntoNotes). These trends are similar for both English and Chinese.

\begin{table}[t]
\begin{center}
\begin{small}
\begin{tabular}{l|lcr}
\toprule
 \textbf{Train} & \multicolumn{3}{c}{\textbf{Test}}   \\
  &\name & \name-Name & $\Delta$\\
\midrule
 ON & 53.41 & 52.92 & -0.49\\
 Friends\textsubscript{CI} & 46.64 & 34.53 & -12.11\\
 \name	& 70.70 & 62.38 & -8.32 \\
 ON+\name & 72.01 & 61.19 & -10.82 \\
 ON$\rightarrow$\name & 75.87 & 73.60 & -2.27 \\\midrule
 \name-Name & 68.91 & 70.88 & \textbf{+1.97}\\
\bottomrule
\end{tabular}
\caption{F1(\%) of models evaluated on original \name~ test set and a version with character names randomly replaced per scene (\name-Name).}
\label{tab:names}
\end{small}
\end{center}
\end{table}

\subsection{Overfitting to Specific Shows}
\label{sec:speaker}
As one of our goals is a dataset enabling a better understanding of multiparty conversations, a concern is that models may overfit to the limited (two) TV shows and the subset of characters (and names) in the training set. While the test set contains our target domain (multiparty conversation), it also shares characters and themes with the training set. 

\paragraph{Names} We test whether models are sensitive to speaker names, perhaps overfitting to the character names and their traits. We replace speaker names in the original \name~dataset with random names. First, we assume the self-identified genders of the speakers through their pronoun usage. Next, for each scene, we replace the name of a character with a randomly sampled name of the same gender.\footnote{We use the top 100 names by frequency for each gender according to \url{https://namecensus.com/}, which is based on the 1990 US Census.} %

The results in Table \ref{tab:names} show that models do overfit to character names: for models trained on \name, Friends\textsubscript{CI}, and ON+\name, performance on \name~test sets drops after replacing names, thereby showing that they are sensitive to names seen in training. On the other hand, both ON and ON$\rightarrow$\name~ show more robustness to changes in speaker name. This is likely because ON does not have a persistent set of characters for the entire training set. It is less clear why ON$\rightarrow$\name~ experiences only a small drop; robustness through continued training can be investigated further in future work. %

We create a \textit{training} set (\name-Name) without a persistent set of characters or speakers by randomly replacing the character names. %
While \name~performance drops slightly compared to a model trained with the original data, it outperforms on the name-replaced test set. Since we have the \{original, replaced\} name mapping, we can convert predictions from \name-Name to \name, resulting in an F1 on \name~competitive with the baseline, after post-processing. These findings support the hypothesis that models that see names used in a ``generic fashion'' are more robust towards name changes \cite{shwartz-etal-2020-grounded}.%

\begin{table}[t]
\begin{center}
\begin{small}
\begin{tabular}{l|lcr}
\toprule
 \textbf{Train} & \multicolumn{3}{c}{\textbf{Test}}   \\
  &\name& \name\textsubscript{Friends}  &\name\textsubscript{TBBT} \\
\midrule
 
 \name	&70.70	&67.02&	71.61\\
 \name\textsubscript{Friends}&	61.95&	61.03&	62.28\\
 \name\textsubscript{TBBT}&	71.22&	69.64&	71.62\\

\bottomrule
\end{tabular}
\caption{F1(\%) of models trained and tested on each of the TV shows, to measure potential overfitting to the training show.}
\label{tab:episode}
\end{small}
\end{center}
\end{table}

\paragraph{TV Series} To determine overfitting to a specific TV show, we split \name~(English) into the two components: \name\textsubscript{Friends} and \name\textsubscript{TBBT}, shown in Table \ref{tab:episode}. %
In this analysis, we find that the variance due to random seed is high, which might explain why training with \name\textsubscript{TBBT} appears to be the best model. %
The results suggest both models find \name\textsubscript{TBBT} easier to predict. Furthermore, training with \name\textsubscript{TBBT} outperforms  \name\textsubscript{Friends} when evaluated on \name\textsubscript{Friends}, suggesting that the substantially larger size of the \name\textsubscript{TBBT} portion beats any in-domain advantages \name\textsubscript{Friends} may have.

\subsection{Alignment Correction}
\label{sec:alignment-correction}
To identify the types of systematic errors made by automatic projection, we analyzed the corrected Chinese alignments. %
Table~\ref{tab:correction} shows the difference in model performance between the corrected and the silver test set. Performance drops a few F1 points on the corrected set, which is caused by the distribution shift from (uncorrected, silver) training data. Naturally, \name-Zh suffers the largest drop because it is closest in the domain to test\textsubscript{silver}. However, it is still one of the best performing models. %

The performance drop of the ON-only trained model is only 0.85 points, possibly because this model is trained on the cleaner (gold) training labels. These observations suggest that while the alignment correction yields a cleaner test set, the automatic silver data is still a good substitute for model development when no gold data is available. 

There is a similar pattern in Farsi. Most of the drop is in recall, since many new mentions are added via alignment correction. These new additions are mostly words that contain compound possessive pronouns or verbal inflectional suffixes that align to a source English word, which are not often captured by automatic word alignment methods. For example, the word 
"\begin{scriptsize}\foreignlanguage{arabic}{\textRL{نميشم}}\end{scriptsize}"
, is a verb with the inflectional suffix "\begin{scriptsize}\foreignlanguage{arabic}{\textRL{ـم}}\end{scriptsize}" aligning to the source mention "I". Another example is "\begin{scriptsize}\foreignlanguage{arabic}{\textRL{دوستت}}\end{scriptsize}", composed of the noun "\begin{scriptsize}\foreignlanguage{arabic}{\textRL{دوست}}\end{scriptsize}" plus the possessive pronoun "\begin{scriptsize}\foreignlanguage{arabic}{\textRL{ـت}}\end{scriptsize}" aligning to the source span "your" in Fig. \ref{fig:projection-example}.

\begin{table}[t]
\begin{center}
\begin{small}
\begin{tabular}{ll|cccccc}
\toprule
 &\textbf{Train} & \multicolumn{3}{c}{\textbf{Test}}   \\
  &&corrected& silver  &$\Delta$ \\
\midrule
 \multirow{7}{*}{Zh} &ON & 36.56 &37.41& 0.85\\
  &ON+\name-Zh & 47.37  &49.38&2.01\\
  &ON$\rightarrow$\name-Zh & 48.65  &51.09&2.44\\
  &\name-Zh & 47.89 &51.87&3.98\\
  &\name-En & 37.18 &39.23& 2.05\\
  &\name-Fa & 37.01 & 39.54& 2.53\\
  &\name-En-Zh-Fa &45.54  &47.30&1.76\\
  \midrule
  \multirow{4}{*}{Fa}&\name-Fa & 35.00 &39.76&4.76\\
  &\name-En & 22.91 &25.07& 2.16\\
  &\name-Zh & 17.09 &20.06 & 2.97\\
  &\name-En-Zh-Fa &33.49  &38.30&4.81\\
\bottomrule
\end{tabular}
\caption{F1 of models on Chinese and Farsi test set before and after correction.}
\label{tab:correction}
\end{small}
\end{center}
\end{table}

\subsection{Annotation from Scratch}
\label{sec:annotation-from-scratch}
Instead of relying on noisy (but free) projections of parallel English data, one could directly annotate coreference in the target language with native speakers. %
To investigate the quality of test\textsubscript{silver} and test\textsubscript{correct}, we perform an analysis study on three randomly sampled scenes from the Chinese test set and ask an annotator to complete the full coreference annotation task. We also obtain oracle word alignments to explore the effect of alignment errors in our data projection framework.

We find MUC score (agreement) rates of 71.84, 78.23, and 87.25 using test\textsubscript{silver}, test\textsubscript{correct}, and oracle projections respectively. This suggests that the corrected test set has a comparable agreement rate to that of the gold data, while the gold projections are also within the range of inter-annotator agreement. As automatic alignment methods improve, our recipe for creating multilingual coreference data will also benefit. Nonetheless, one of the limitations of \name~is that quality of the Chinese and Farsi test sets could still be higher. %

\paragraph{Advantages}
Despite lower quality, the data projection method still has several advantages over from-scratch annotation as it is faster and there is less demand for an in-language expert. 

First, annotation from scratch requires a syntactic parser to find constituencies for mention linking (Sec. \ref{sec:english}). The zero-shot transfer setting usually involves lower-resource languages, where parsers, if they exist at all, may not perform well. %
Thus, projection may be the only solution in these cases.

Second, the annotation quality depends on the guidelines. Linguistic experts in the target language will need to design annotation guidelines and experts are not always available. However, this step can be skipped with projection (since we are releasing \name, which has parallel text in numerous languages). Not only the projection task itself is significantly simpler to explain, it is easier to understand and can be faster than annotating from scratch. %
In our setting, around 70\% of the predicted alignments were marked as correct. One could design heuristics to only present the difficult mention pairs, which would further reduce annotation cost.

\section{Conclusion}

Motivated by a desire to better understand spontaneous multilingual conversations, we developed a collection of coreference annotations on the transcripts and subtitles of two popular TV sitcoms. To reduce the cost of annotating from scratch for each language, we selected our English data such that there were already existing gold human translations available in the form of subtitles, in order to automatically project our annotations from English. After manually correcting these projections, 
we observe a few point differences in reported values across various multilingual models.

There exist dozens of additional languages that our annotations may be projected to in the future. If automatic projection leads to only a few point variance in the estimated performance of a model, we believe this framework is sufficient for driving significant new work in coreference across many non-English languages in the future.

\section{Limitations}
There are several limitations in the dataset inherent to the difficulty of the task, crowdsourcing, and the use of models for candidate proposals. The inter-annotator agreement scores are not perfect. One contributing factor is that we do not post-process or provide explicit instructions for pleonastic pronouns, so annotators used their own judgment. These account for 3.15\% of the mentions in the pilot annotation. There is also a distribution difference between the (noisier) train and dev/test set caused by different annotator sources, how they were paid, and whether the annotations were adjudicated. Additionally, annotation was performed without access to ground truth video, which could impede annotation or encourage guessing when situatedness may be required. Since annotation in MMC is aided by other models (parser and aligner), system errors may not necessarily be caught during annotation.

\appendix
\section*{Appendices}
\section{Split Antecedent Statistics}
\label{sec:appendix-drop-split-statistics}
\name-En has a number of split antecedents; 1,156 antecedents across 2,745 spans in the training set; 255 across 717 spans in the dev set, and 178 across 444 spans in the test set.

\section{Merging Two-way Annotations}
\label{sec:merge-two-way}
A third annotator adjudicates disagreement in the two-way annotations in the dev/test set. %
To decide whether a pair of annotations disagree, we first build common clusters between two annotations. After annotation, each query mention is annotated with two antecedents.  $$A = \{(q_1, a_1^{1}, a_1^{2}), (q_2, a_2^1, a_2^{2}), ..., (q_n, a_n^1, a_n^{2})\}$$
$q_i$ is the $i^{th}$ query and $n$ is the number of candidate queries. $a_i^{1}$ and $a_i^{2}$ are the antecedents linked to the $i^{th}$ query. We build initially agreed clusters by taking the transitive closure of the subset of A where each triplet agrees exactly (i.e. for $q_i$, $a_i^{1} = a_i^{2}$) between the two annotations. Note that the annotations, $a_i$, can be another query span, $q_j$, that is also annotated. This lets us connect the annotations and form clusters.

Next, we incrementally add query spans to these clusters if both annotators link them to the same cluster ($a_i^{1} \neq a_i^{2}$ but $a_i^{1}$ and $a_i^{2}$ are in the same cluster anyway), continuing until no further pairs agree. At the end, if there exist $q_i$ where $a_i^{1} \neq a_i^{2}$, then each $(q_i, a_i^{1}, a_i^{2})$ is marked for adjudication. The adjudicator is prompted to select between $a_i^{1}$, $a_i^{2}$, or relabel $q_i$ entirely. Their annotation is final.

\section{Word Alignment}
\label{sec:appendix-awesome}
Word alignments are extracted from the fine-tuned XLM-R-large model using Awesome-align. We first fine-tuned XLM-R on English-\{Chinese, Farsi\} parallel data that has been filtered using LASER semantic similarity scores \cite{schwenk-douze-2017-learning, thompson-post-2020-automatic}. We reuse empirically-chosen Awesome-align hyperparameters from prior work for a similar task \cite{yarmohammadi-etal-2021-everything}: softmax normalization with probability thresholding of 0.001, 4 gradient accumulation steps, 1 training epoch with a learning rate of $2 \cdot 10^{-5}$, alignment layer of 16, and masked language modeling (``mlm''), translation language modeling (``tlm''), self-training objective (``so''), and parallel sentence identification (``psi'') training objectives. We further fine-tuned the resulting model on the gold word alignments on 1500 En-Fa and 2800 En-Zh sentence pairs with the same hyperparameters, for 5 training epochs with a learning rate of $10^{-4}$ and only ``so'' as the training objective.

\section{Hyperparameters}
\label{sec:appendix-hyperparameters}

We reuse most of the hyperparameters from \citet{xu-choi-2020-revealing}: we enumerate spans up to a maximum span width of 30 and set the maximum speakers to 200, ``top span ratio'' to 0.4, and maximum top antecedents (beam size) to 50. For XLM-R models, we set the LM learning rate to $10^{-5}$ and task learning rate to $3\cdot 10^{-4}$. For SpanBERT models, we use a LM learning rate of $2\cdot 10^{-5}$ and task learning rate of $2\cdot 10^{-4}$.

Following a grid search, we set the mention loss weights $(\alpha_m)$ for the each language and dataset: 5 for \name-Zh and \name-En, 6.5 for \name-Fa, and 0 for OntoNotes. For $\tau$ we find $\tau_{\mathrm{Fa}}=0.55$, $\tau_{\mathrm{Zh}}=0.7$, and $\tau_{\mathrm{En}}=0.7$ performed best on dev.

\section*{Acknowledgements}
We thank Michelle Fashandi and other linguists at HLTCOE, along with Chenyu Zhang and Kate Sanders, for their annotation efforts. We thank Elias Stengel-Eskin, Kate Sanders, and Shabnam Behzad for helpful comments and feedback and Chandler May for help in building the annotation interface. The third author acknowledges support through a fellowship from JHU + Amazon Initiative for Interactive AI (AI2AI). This work was supported by DARPA AIDA (FA8750-18-2-0015) and IARPA BETTER (201919051600005).
The views and conclusions contained in this work are those of the authors and should not be interpreted as necessarily representing the official policies, either expressed or implied, or endorsements of DARPA or the U.S. Government. The U.S. Government is authorized to reproduce and distribute reprints for governmental purposes notwithstanding any copyright annotation therein.

\bibliography{anthology,custom}
\bibliographystyle{acl_natbib}

\end{document}